# Image Processing on IOPA Radiographs: A comprehensive case study on Apical Periodontitis

Diganta Misra, Vanshika Arora

*Abstract*— With the recent advancements in Image Processing Techniques and development of new robust computer vision algorithms, new areas of research within Medical Diagnosis and Biomedical Engineering are picking up pace. This paper provides a comprehensive in-depth case study of Image Processing, Feature Extraction and Analysis of Apical Periodontitis diagnostic cases in IOPA (Intra Oral Peri-Apical) Radiographs, a common case in oral diagnostic pipeline. This paper provides a detailed analytical approach towards improving the diagnostic procedure with improved and faster results with higher accuracy targeting to eliminate True Negative and False Positive cases.

## I. INTRODUCTION

THE medical diagnostic pipeline is a crucial procedure underlying the following steps taken based on the results obtained from diagnosis. Medical Experts and Doctors rely on Pathological and Diagnostic Labs to take subsequent actions. One of the most common instances of diagnosis is based on X-Rays and Radiographs, along with Resonance Imaging. These techniques are highly crucial in understanding and verifying the case involved. Mostly each and every medical domain involves imaging techniques to study the patient ranging from fMRI, MRI, CT-Scans, X-Rays and general Radiographs; Image Processing techniques come into play and have very high importance in this procedure to enhance results, increase robustness and make the process faster. However, the most crucial aspect of it is to remove cases of False Positives and True Negatives. Due to these two scenarios, some patients are misclassified to be healthy and vice versa.

Image Processing [1] is the cumulative term for the diversified variety of types of techniques applied to digital images to maximize information gain, enhance image quality, to advance analysis and to recognize underlying patterns for feature extraction and analysis. Digital Image Processing has been at the forefront in Biomedical Image Analysis not only due to the ability to improve results but in general to enhance image analysis and produce more robust and high efficient algorithms. Medical Science domains demand a high level of precision and accuracy when it comes to image analysis especially of radiographs or resonance imaging results, and as a result the need of a highly efficient and robust Image Processing Pipeline is highly necessary to improve diagnostic procedures.

Any image which is generated on a sensor plate or film by any rays of particular wavelength, usually X or gamma rays or comparable radiation is known as a radiograph [2]. The beams enter the oral structures relying on the distinction in their anatomical introduction and forces. Teeth are typically observed lighter or murky in the radiograph as a result of the absence of penetrating rays that hit them. So if, there is any nearness of pathologies for instance dental caries or variety in bone thickness in the structure, the district can appear darker or translucent because then the x-rays that enter through can also pass through the abnormal structure. Dental fillings, restoration, implants and other intra oral machines have diverse yield on the radiograph relying upon their material thickness. A dental radiograph is normally arranged into:

a.) Intra Oral Radiographs-The radiographs which are taken by putting the film inside the patient's mouth and afterward beginning the radiation are called intra oral radiographs. The subclasses incorporate periapical, bitewing, occlusal and full mouth arrangement.

b.) Extra Oral Radiographs-These radiographs are taken by setting the sensor or the film outside the oral hole of the body. These generally incorporate Panoramic movies [3] [4].

Intra oral periapical radiography is a normally utilized intraoral imaging system in dental radiology and might be a segment of intraoral periapical radiologic examination. Periapical radiographs give vital data about the teeth and adjacent bone structure. The X-ray taken shows the whole crown and foundation of the teeth and the adjacent alveolar bone which gives fundamental data to help in the determination of the most widely recognized dental lesions and diseases ; particularly dental caries, tooth abscesses and periodontal bone misfortune or gum ailment. Extra vital discoveries might be identified, including the state of rebuilding efforts, affected teeth or broken tooth parts and varieties in tooth and bone structures [5]. Digital radiography does not remain a trial methodology anymore. It is a dependable and flexible innovation that expands the diagnostic and image sharing potential outcomes of radiography in dentistry [6]. Proficiency in understanding radiographs and finding abnormalities is achieved by sufficient amount of diagnostic image exposure, search strategies for visual interpretation, being able to able to identify and reason any abnormality in oral radiograph [7].

Diganta Misra is with the School of Electronics Engineering, Kalinga Institute of Industrial Technology (KIIT) Bhubaneswar, Odisha 751024 India (corresponding author, e-mail: mishradiganta91@gmail.com).

Vanshika Arora is with the Dental Science Department, Manav Rachna Dental College, Faridabad, Harayana 121004, India (e-mail: vanshikaarora25@gmail.com).

## II. RELATED WORK

Digital image analysis in radiology [8] has turned out to be one of the standard aptitudes of technologists and radiologists alike. It is imperative that technologists comprehend the nature and degree of computerized pictures as well as of advanced image processing, to wind up viable and effective clients of the new innovations that have had a huge effect on the consideration and the board of patients.

The real objective of advanced picture post-processing in therapeutic imaging is to adjust or change a picture to improve analytic elucidation. For instance, pictures can be post-processed with the end goal of picture upgrade, picture rebuilding, picture investigation, and picture pressure. These tasks are expected to change an info picture into a yield picture to suit the review needs of the spectator in making a finding.

In [9] paper, researchers have shown how image processing techniques can be helpful for not only the detection of dental caries on radiographs but also classifying their extent and site. Apart from diagnostics, image processing is also applicable in prosthetics and implantology as mentioned in paper [10] by evaluating density and grey-scale changes in the bone pattern on radiographic examination processes which provide various information about the quality, measure, site and width of the bone, the occlusal pattern of the teeth and also for diagnosing any odontogenic tumors or lesions. Thus, confirming the number and the type and size of implants to be used by the clinician. In paper [11], the authors have focused on the endodontic aspects of this application where they presented a comprehensive evaluation of a Root Canal Treatment to propose the length of the root canal semi-automatically.

In [12], the author has summarized the fundamentals of biomedical image processing. The paper has explained the core steps of image analysis which includes feature extraction, segmentation, classification, quantitative measurements, and interpretation in separate sections. In [13], the authors have devised an extensible, platform-independent, general-purpose image processing and visualization program called MIPAV (Medical Image Processing, Analysis and Visualization) which was specifically designed to meet the needs of an Internet-linked medical research community. The authors in [14] have proposed a novel approach for Biomedical Image Segmentation based on a modified version of Convolution Neural Networks known as U-Nets which rely on Data Augmentation for end-to-end training. In [15], the paper is essentially a critical review which talks about how the last two decades have witnessed a revolutionary development in the very field of biomedical and diagnostic imaging.

## III. PERIODONTITIS

By and large, as a result of dental caries or poor oral wellbeing, plaque, tartar and calculus start accumulating around the teeth and gingiva making the gingiva swollen, fragile and bright pink. This condition is frequently named as gum infection or gum disease preferably known as gingivitis. Exactly when this gum disease is left long-standing and untreated, this may change into periodontitis [16]. An irritation connecting past the gingiva pounding the connective space around the tooth is named as apical periodontitis. In this condition, gums will in general draw a long way from teeth making little spaces which harbor the advancement of micro-organisms are named as "periodontal pockets". To control the advancement of these species, our very own body's safe response turns out to be perhaps the most essential factor as it fights the microscopic creatures which have now started to create underneath the gum line (cervico-gingival line). Bacterial toxic substances and body's invulnerable structure, on account of their movement, start to invade the bone and connective tissue which help in holding the dentition. So if not treated, the bones, gums and the including tissues are demolished at last.

This can much of the time incite transportability of teeth due to removing, birth of more bothers including boil or granuloma and we would need to trade off the tooth [17]. This infection has been characterized into two
Sub-classes, constant, forceful or because of any indication of fundamental disorder [18].

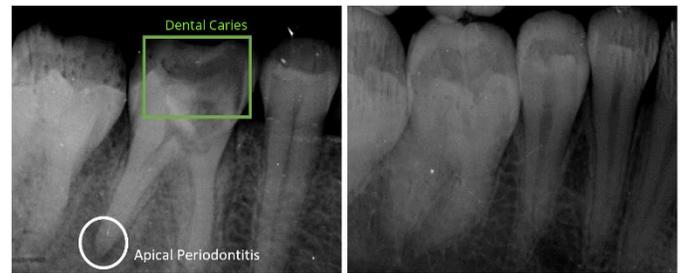

Fig. 1 (a). Dental Caries affected Mandibular Left First Molar approaching the pulp disto-occlusally with evident PDL widening in the distal root suggestive of apical periodontitis. (b). Healthy Molar with same region of interest. (Image Credits: Department of Oral Radiology and Medicine, Manav Rachna Dental College)

Fig. 1 shows the comparison between an affected mandibular molar with evident dental caries and PDL (Periodontal Ligament) widening in the distal root, a clear indication of the case of Apical Periodontitis and a healthy mandibular molar. The affected sample radiograph is further used in the paper for Image Analysis and several image processing techniques mentioned in subsequent sections in the paper are applied for further inferences and observations.

### A. Chronic Periodontitis

It is considered as the most notable frame which has a moderate advancement rate and is typically found in the elderly yet much of the time can be found in youths as well. Early signs of this ailment are commonly left unnoticed which may consolidate halitosis (horrendous breath), redness and bleeding gums which for the most part happens while brushing or eating anything hard in consistency, gum swelling that keep rehashing, receding gums (this happens due to the subsidence of gingiva from the cervico-gingival edge) and it hangs tight, might provoke loosening of teeth as well. Regardless of the way that the development is moderate on a typical, the patient may keep running over occasions of sudden 'impacts' or quick development which can be agonizing. Interminable periodontitis has consistently been associated with smoking, nonappearance of oral neatness with insufficient plaque control or by principal factors, for instance, HIV(Human immunodeficiency virus) infection or Diabetes Mellitus [19].

*B. Acute Apical Periodontitis*

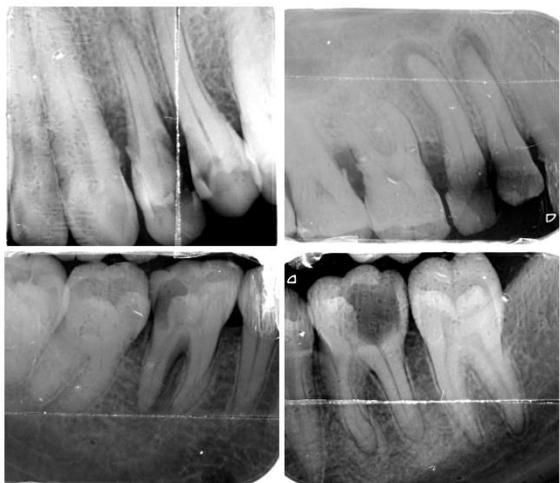

Fig. 2 (a). Dental Caries affecting maxillary first premolar from the left from the distal aspect with initiation of PDL widening suggestive of apical periodontitis. (b). Dental Caries affecting the maxillary right premolars with prominent periodontal ligament widening around the affected roots suggestive of apical periodontitis. (c). Dental Caries affecting mandibular right first molar with PDL widening prominently seen over mesial root suggestive of apical periodontitis. (d). Dental caries affecting left mandibular second molar where the caries has invaded the pulp completely from the distal aspect with PDL widening in both roots indicating chronic apical periodontitis. (Image Credits: Department of Oral Radiology and Medicine, Manav Rachna Dental College)

Usually, when any kind of degradation or infection approaches the periodontal ligament after exiting from the pulp, it causes inflammation. Even though in such cases there can be no visible abscess seen, but can be diagnosed with the help of IOPAs. Acute apical periodontitis is usually caused by trauma or wedging. Traumas are usually by any defect in occlusal contact or any undue pressure caused by the same. Traumas are also very common when any restoration is faulty or is inserted more than required. Wedging as the name suggests happens because of any lodgement of any foreign bodies which cause inflammation. This inflammation is usually limited to the periodontal ligament and the patient feels sharp throbbing pain especially on percussion [20].

Fig. 2 is a collection of variously affected sample radiographs showing prominent Dental Caries and Apical Periodontitis both in the mandibular and maxillary premolars and molars. The extent of the same can be visually interpreted from the given radiographs.

## IV. IMAGE PROCESSING TECHNIQUES

Image processing [1] is the collective term given to the set of techniques which can be applied to images on a digital perspective which refers to techniques or algorithms including Image Enhancement Techniques, Edge Detection, Feature Extraction, Pattern Recognition, Contour Modelling, Entropy and Magnitude/Phase Spectrum Analysis, Rescaling, Localization, Histogram Equalization, et cetera. All these techniques maximize information gained from the image. Image Processing is very similar to signal dispensation where it takes an image to be the input and can produce an output either to be an enhanced image or characteristics or information extracted from that image. Image Processing is a rapidly growing domain with increasing demand in almost all commercial and academic sectors including Agriculture, Astronomy, Medical Science, Surveillance, Autonomous Modes of Transportation, Photography, et cetera.

Digital Image Processing especially in the medical science domain is a crucial underlying procedure especially when dealing with digital radiographs [21] and this paper is constructed in consideration to the high level of impact of Image Processing in the diagnostic pipeline for advanced clinical research.

*A. Basic Image Transformation*

Initial Image Processing Techniques involve basic image transformations which include rotational changes on both vertical and horizontal axes, scaling of an image and also localization of object of interest in the image. This is normally done to find out the factor of spatial variance/ invariance of the orientation of the image and also to determine focal points or central co-ordinates of the image. Some Object Segmentation techniques are sensitive to spatial orientation of the object in the image or are even scale variant and this process helps in benchmarking these segmentation techniques.

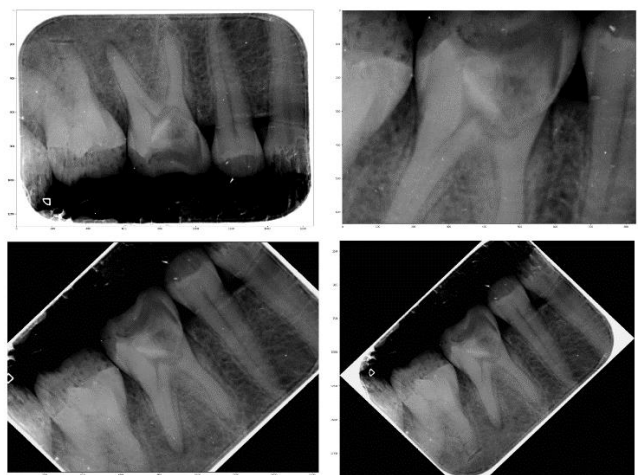

Fig. 3 (a). Vertical Flip of the test image (b). Zoomed and Rescaled Test Image. (c). Non-Reshaped Test Image Rotated 45° anti-clockwise. (d) Reshaped Test Image Rotated 45° anti-clockwise.

Fig. 3 shows some basic transformation applied on the test image including Flipping on the vertical axis, Rescaling and Zooming of the Test Image with localization on the Dental Caries Region of Interest (RoI) on the affected Molar and Reshaped along with Non-Reshaped anti-clockwise rotated test image.

Fig. 4 shows the central co-ordinates of the image which was found out to be (732.2945246039354, 717.4333518192004) based on the actual resolution of the image which is 1657×1270 pixels with both horizontal and vertical resolution of 1011 dpi (dots per inch) and bit depth of 8.

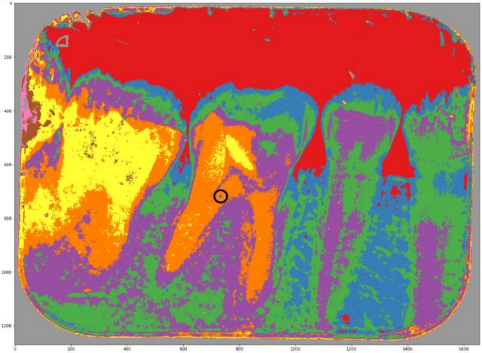

Fig. 4. Central Co-ordinates of the input test image in *Set1* color-map.

*B. Image Blurring, Noise Removal and Smoothening*

An important procedure in Image Processing involves blurring or smoothening the image primarily focused on reducing noise levels while trying to preserve the edges and information present within the image.

*1) Gaussian Blurring:*

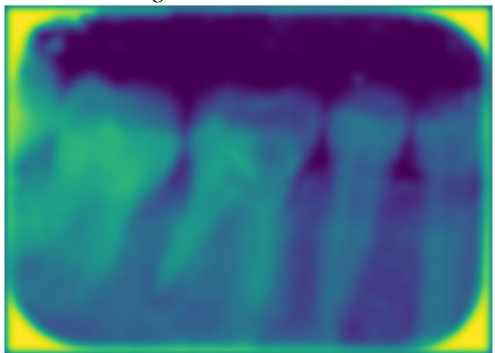

Fig. 5. Gaussian Blurring applied on single channel reduced image.

Gaussian Blur [22]-[23] is a very commonly applied type of blurring technique inspired from the standard Gaussian Distribution which is defined in terms of a mathematical equation to be as:

$$P(x) = \frac{1}{\sigma\sqrt{2\pi}} e^{\frac{-(x-\mu)^2}{2\sigma^2}} \quad (1)$$

σ represents the standard deviation while μ represents the mean of the distribution A kernel is built using this equation and applied on the input image to obtain the blurred output. Fig. 5 shows the output of a standard Gaussian Blurring Kernel [23] applied to the test radiograph input. Gaussian distribution is shown in both 2-dimensional and 3-dimensional aspects in Fig. 6.

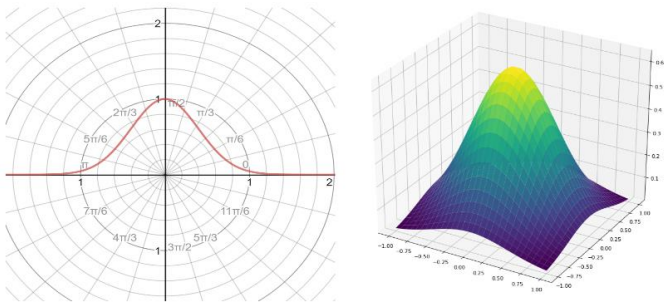

Fig. 6. (a). A 2-D centered uniform Gaussian distribution. (b). 3-D representation of a standard Gaussian distribution. [22]

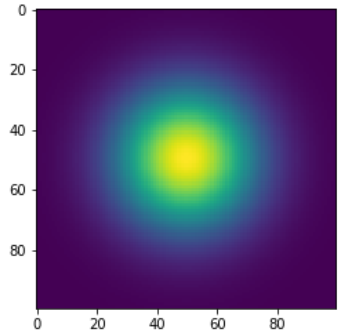

Fig. 7. A standard Gaussian Blurring Kernel [22]

Fig. 7 shows the standard Gaussian Blurring kernel which has been applied on the test input radiograph to obtain the blurred image as shown in Fig. 5 with reduced noise levels including decreased sharpness and details.

*2) Median Blurring:*

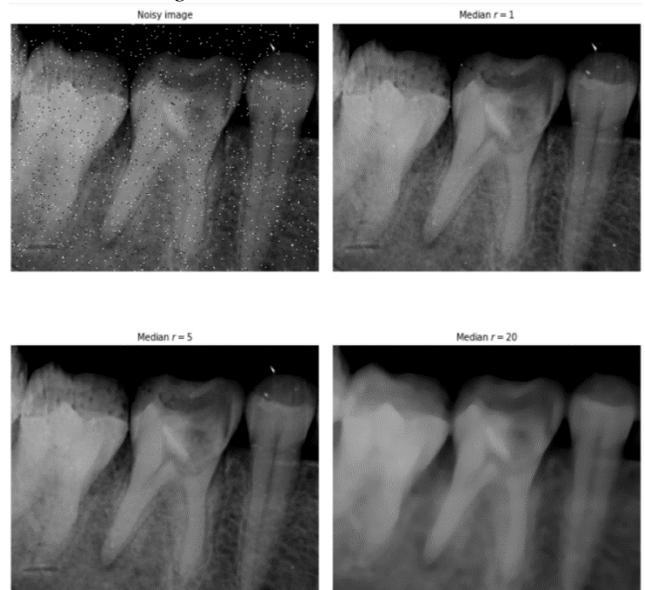

Fig. 8. (a). Input Test Radiograph with modulated Salt and Pepper noise (b). Median Blurring with r = 1px applied on the noisy image. (c). Median Blurring with r = 5px applied on the noisy image. (d). Median Blurring with r = 20px applied on the noisy image.

Median Blurring [24] is a common technique for blurring and reduction of noise levels in an input image suffering from high magnitudes of corrupted pixels. This technique is popular as it helps to process the image and enhance its quality for later processing stages and over that it also preserves the edges of the image thus helping in contour modelling or edge detection.

Median Blurring [24] follows the simple algorithmic procedure of replacing a target pixel entry with the median of the neighboring pixels defined by the boundary set by the radius specified by following a sliding window pattern. The popularity of median blurring is immense due to its efficiency in removing random noise patterns while preserving the original orientation and information of the image as compared to Gaussian Blurring where the image is blurred with a trade-off to losing the sharp edges in the image. In Fig.8, we have added random salt and pepper noise to the input test radiograph and tested the efficiency of standard median blurring with 3 different radius values: 1px, 5px and 20px. As shown Median Blurring was successfully capable in removing the noise pattern while preserving the original image information and it is also shown as the radius increases, the blurring factor increases thus defining and confirming the direct relationship between the two. Median Blurring can be highly beneficial in Digital Radiograph analysis especially in Apical Periodontitis case of IOPA (Intra-Oral Per-Apical) due to bad radiograph quality or distortion of pixel intensity values in the gum region of the oral cavity which may allow the pathology to be camouflaged within the tissue layer shown in the radiograph and lead to incorrect diagnosis.

*3) Local Mean Blurring and Bilateral Mean Filtering:*

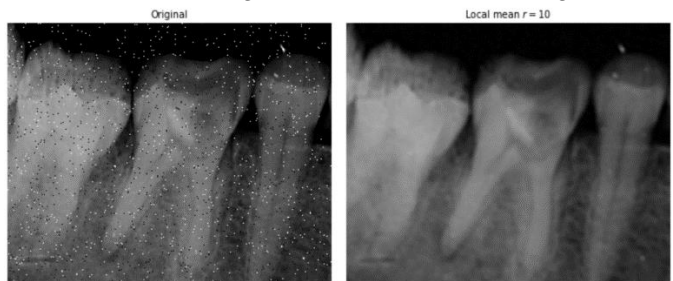

Fig. 9. (a). Input Test Radiograph with modulated Salt and Pepper noise (b). Mean Blurring with r = 10px applied on the noisy image.

Like Median Filtering, Local Mean Blurring[24]-[25] has the same functionality but uses mean or summed average rather than median of the neighboring pixel for the target pixel entry within the boundary defined by the radius value taken for the blurring algorithm. Fig. 9 shows the application of local mean blurring on the same input test radiograph with modulated salt and pepper noise and as it can be seen, the filter is capable of removing the noise while preserving the edges present in the image.

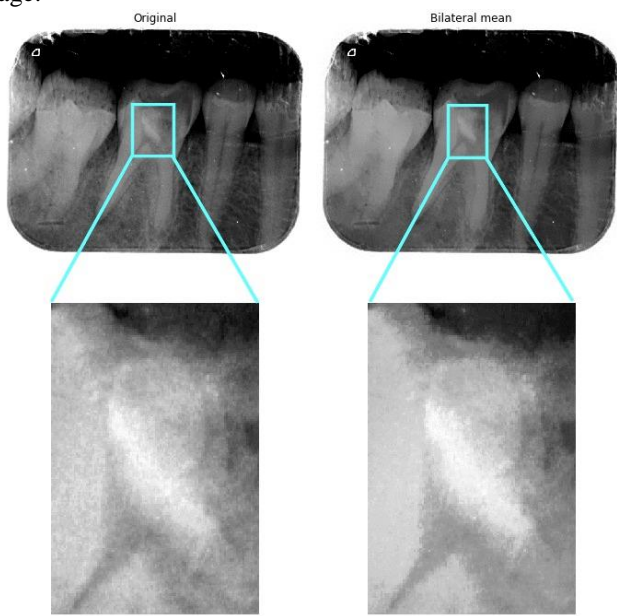

Fig. 10. (a). Input Non-Modulated Test Radiograph (b). Bilateral Mean Filtering applied on the Input Radiograph.

Bilateral Mean Filtering [25] is a non-linear image smoothening filter which just as its predecessor serves the purpose of removing noise from the image while preserving the edges. However, it follows the algorithmic procedure of replacing the target pixel with a weighted average of the pixel intensities neighboring pixels. The weights used for calculating the average maybe derived from a centric uniform distribution just like Gaussian distribution as defined in previous section of this paper. Primarily the weights don't only have the Euclidean distance as the dependency but also radiometric differences like depth differences, varying color intensity, et cetera which in return helps to preserve sharp edges present within the image.

The Bilateral Mean Filter can be mathematically defined as:
$$I^{output}(x) = \frac{1}{W_p}\sum_{x_i\in\Omega} I(x_i)f_r(||I(x_i) - I(x)||\ )g_s(||x_i - x||) \quad (2)$$

And the normalization term to be:
$$W_p = \sum_{x_i\in\Omega} I(x_i)f_r(||I(x_i) - I(x)||)g_s(||x_i - x||) \quad (3)$$

Where $I^{output}$ is the output filtered image after applying the Bilateral Mean Filter on the input image defined as $I$. $x$ are the co-ordinates of the target pixel which is to be filtered, $\Omega$ is the window centered on the target pixel, $f_r$ is the range kernel which is applied for smoothening intensity differences and $g_s$ is the spatial kernel for smoothening co-ordinate differences.

Fig. 10 shows Bilateral Mean Filtering applied on the test input radiograph and as shown the noise level was considerably decreased as well as the image was smoothened while preserving the sharp edges and might prove to be a great alternative to other conventional techniques of Image Blurring and Smoothening

*C. Morphological Image Processing*

Morphological Image Processing [26] is the section of Image Processing comprising of a set of non-linear processing techniques involving the shape or the morphological orientation of the features in the image which is defined to rely on the relative ordering of the pixel values only.

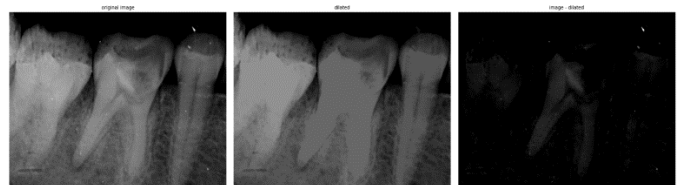

Fig. 11. (a). Original Image (b). Dilated Image (c). Image Subtracted from the Original Image

Morphological Image Processing [26] involves some basic and some compound operations, one of them including Dilation. Dilation [27] is the process of replacing the target/ structural pixel with the maximum of all the pixels in its neighborhood which is just the opposite of erosion where it takes the minimum instead of the maximum. Dilating an image results in the addition of a layer of pixels to both the inner and outer boundaries of a region in the image. Dilation of an image *f* by a structural element *s* is denoted as:
$$f \otimes s \quad (4)$$
Dilation is used for increasing the brightness of the objects present in the image and is also used to fill gaps/ holes present in the image. Fig. 11 shows image dilation applied on the original image and later the image is subtracted from the dilated image to observe the effects of dilation on the image. Further in Fig. 12, we see the effect of dilation on a slice taken at y = 197 of the input image.

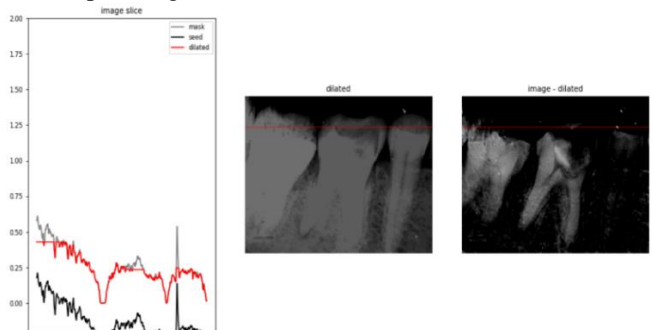

Fig. 12. (a). Intensity Difference Values across the slice y = 197 (b). Dilated Image. (c). Image Subtracted from the Dilated Image.

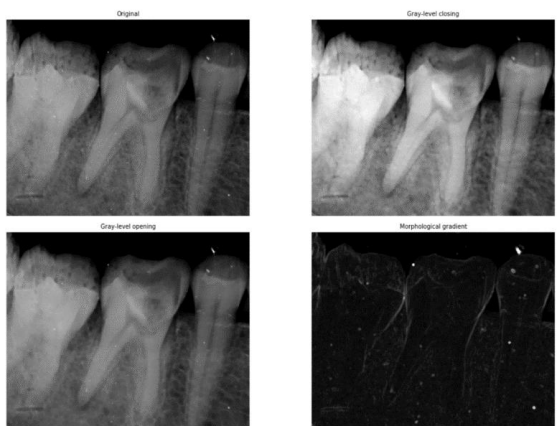

Fig. 13. (a). Original Image (b). Gray-Level Opening Applied on the Image (c). Gray Level Closing Applied on the Image. (d). Morphological Gradient of the Image

Fig. 13 shows the compound operations of Morphological Image Processing which includes: Opening, Closing and Morphological Gradients.

*1) Opening:*

Opening [28] an image *f* means to apply dilation on an image after it has already been eroded, and can be represented mathematically to be:

$$f \circ s = (f \ominus s) \oplus s \qquad (5)$$

, where *s* is the structural element being used. Opening is an idempotent technique which means once applied to an image, subsequent openings with the same structural element won't have any effect. Opening is used to open up the gaps which are bound by narrow layer of pixels. Opening of an image can be very useful especially in analysis of radiographs as it is less destructive than erosion but also tends to reduce the intensity of any features smaller than the structural element while do the exact opposite with the features bigger than the structural element which helps in preserving certain foreground areas of the image which can contain the structural element while attenuating other foreground regions and thus can help in noise removal, better representation of features and also in removal of texture fluctuations which might cause the minute diagnostic regions to blend in with the irregular texture pattern present in the image. As seen in Fig. 13, the gray-level opening of the image causes it to have a more matt look, where the texture has been smoothened and the larger features are now more bright.

*2) Closing:*

Closing [29] of an image is similar to opening and can be derived from the basic mathematical morphological operations of Erosion and Dilation. Closing of an image *f* means to apply erosion on an already dilated image, however the dilation and erosion are performed by 180° rotation of the structural element *s* and can be mathematically denoted as:

$$f \bullet s = (f \oplus s_{rot}) \ominus s_{rot} \qquad (6)$$

, where $s_{rot}$ is the rotated structural element taken. It has the clear opposite effect to that of Opening by which it closes up holes or tiny gaps between regions in the image, and is an idempotent technique just like the Opening of an image is. Closing of an image is used to preserve background regions which can contain the structural element and eliminate the remaining background regions. It also helps in noise reduction, removing irregularity in texture layout and also to remove extreme contrasting features to produce a more smoothened output. As it can be seen in Fig. 13, we can see that the dark irregular patterns in the gum tissue region have been greatly reduced and extreme high contrasting spots are now dampened in intensity.

*3) Morphological Gradient:*

Morphological Gradient [30] of an Image is the difference between the dilation and the erosion of that image and is used to find differences in contrast intensities which is extremely helpful in edge detection and also for object segmentation. Mathematically, it can be denoted as:

$$G(f) = f \oplus b - f \ominus b \qquad (7)$$

, where *f* is the grayscale image and *b* is a symmetric gray-scale structuring element having a short-support. As it is shown in Fig. 13, the Morphological gradient of the input test radiograph has clearly defined the edges and boundaries present in the image and this can be extremely helpful in bounding regions of interest in diagnostic reports and even locate irregular bone structure, crown decay and peri-apical cyst cases in dental pathological IOPA radiographs.

*D. Pixel Intensities*

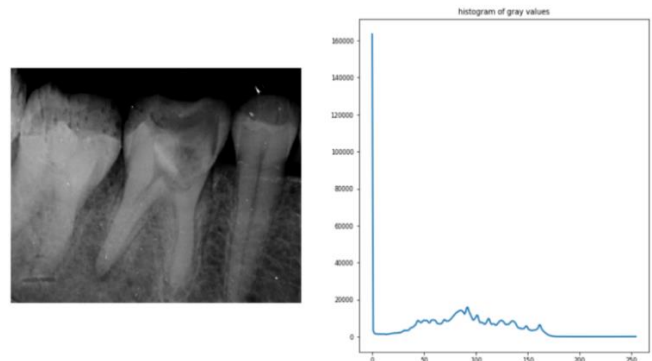

Fig. 14. (a). Original Image (b). Histogram of Gray-scale pixel intensity values

Pixel represents the individual building blocks of an image and has a magnitude which is called the Pixel Intensity which represents the grayscale intensity at that portion. Although rather basic, understanding Pixel Intensities and the variation of intensity in the image can be extremely helpful especially when dealing with biomedical radiographs as it can help you interpret irregularity in the image due to high/ low contrasts which are defined by the Pixel Intensities. Fig 14 shows the histogram of gray-values while Fig. 15 shows the pixel intensity on a zoomed region of interest in the radiograph taken.

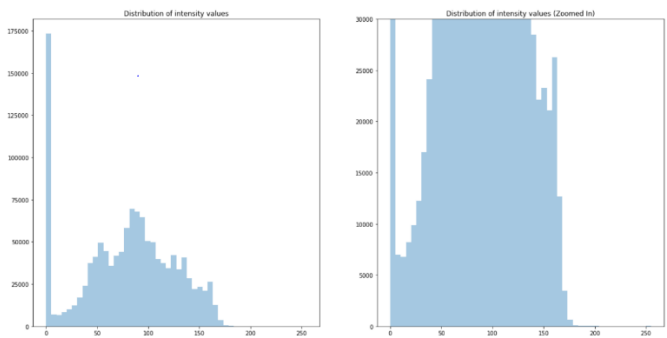

Fig. 15 (a). Mapping of Pixel Intensity values (b) Distribution of grayscale intensity values on the zoomed region of interest of the image.

*E. Feature Descriptors, Feature Matching, Feature Extraction and Analysis*

Undoubtedly one of the most important processes in the Image Processing pipeline, feature extraction [31] and analysis boasts huge importance in helping to understand the image better. Feature Extraction [31] and Analysis many sublevel techniques including application of various Feature Descriptors to gain more insight into the image data for pattern recognition. In the subsequent sub-sections various Feature Extraction and analysis techniques are discussed in detail pertaining to its use cases in dental diagnosis taking the instance of the Apical Periodontitis radiograph sample.

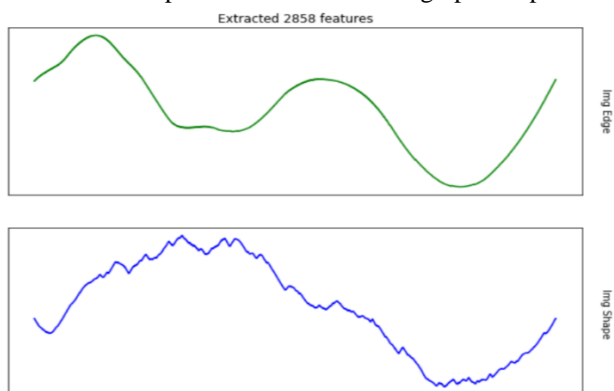

Fig. 16 Extracted 2858 features from the sample IOPA radiograph and the variations in Image Edge and Image Shape.

Upon Initial Analysis we found out 2858 feature key-points on the sample IOPA radiograph which are computed on the basis of areas of high cosine similarities in texture patterns, variation in foreground and background in regions, et cetera. The variation in the Image Shape and Image Size with respect to the image's Cartesian contour mapping is shown in Fig 16.

*1) Histogram of Oriented Gradients (HOG):*

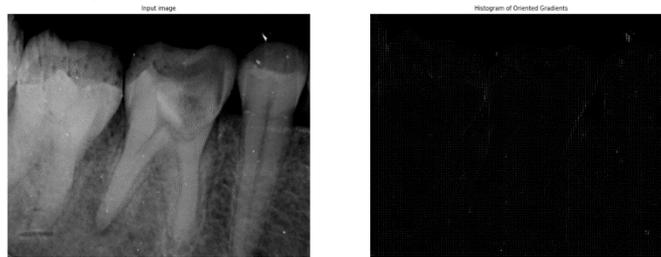

Fig. 17 (a). Original Sample Radiograph (b) Histogram of Oriented Gradients (HOG).

A popular feature descriptor commonly used in Computer Vision and Image Processing, Histogram of Oriented Gradients (HOG) [32] is a feature descriptor used for object detection present in the image. It is computed on a dense grid of uniformly aligned and spaced cells and involves overlapping local contrast normalization to improve accuracy. The whole algorithmic application can be divided into sub-steps which defined in a chronological order of execution are as follows: Gradient Computation, Orientation Binning, defining Descriptor Blocks followed by Normalizing those blocks. In Fig. 17, we can see the HOG of the input sample radiograph, where the edges are clearly defined by descriptor blocks, the individual tooth has been well segmented from each other while also maintaining crown patterns of the tooth which can be extremely useful in detecting Dental Caries, a pathological case in Oral Diagnosis.

*2) Blob Detection:*

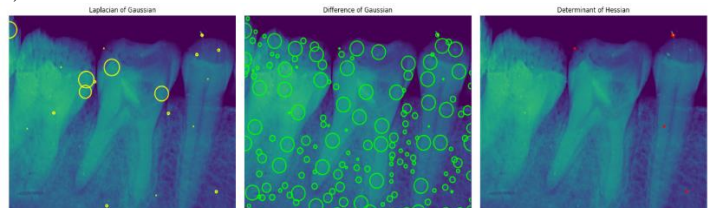

Fig. 18 (a).Laplacian of Gaussian Blob Detection. (b). Difference of Gaussian Blob Detection. (c).Determinant of Hessian Blob Detection.

Blob Detection [33] is an extensive procedure in computer vision and image processing used to understand regions defined as blobs within the image which have stark contrasting features in respect to other blobs which means an area within a blob is somewhat synonymous and similar in perspective of various factors including intensity, color, et cetera. Blob Detection can be extremely helpful in medical diagnosis to provide supplementary information about regions in the image which conventional edge detectors or segmentation algorithms fail to provide and can help in finding irregularity and analyzing affected regions in diagnostic radiographs. Blob Detection has primarily 3 procedures including: Laplacian of Gaussian, Difference of Gaussian and Determinant of Hessian. Blob Detection follows an important underlying mathematical operation fundamental to all types of Blob Detector which is known as Convolution.

Laplacian of Gaussian (LoG) [34] is one of the most common types of Blob Detection where an input image denoted as *f(x,y)* is first convoluted with a Gaussian kernel which is defined to be as

$$g(x,y,t) = \frac{1}{2\pi t} e^{-\frac{x^2+y^2}{2t}} \quad (8)$$

, where $t$ is the scale space factor. Then the convolution will be represented as:

$$f(x,y) * g(x,y,t) \quad (9)$$

The result of the Laplacian then would be:

$$\nabla^2 L = L_{xx} + L_{yy} \quad (10)$$

which gives a high positive response for dark blobs of radius $r = \sqrt{2t}$ and high negative response for bright blobs with the same radius. However, to avoid its shortcomings, dynamic $t$ is taken. Fig. 18 shows the Laplacian of Gaussian Blob Detector applied on the sample radiograph.

Difference of Gaussian (DoG) [35], a very similar approximation of LoG takes the difference between two Gaussian Smoothened Images where the blobs are detected from the scale-space extrema of the difference of Gaussians. Mathematically, the DoG algorithm can be represented as:

$$\nabla^2_{norm} L(x,y;t) \approx \frac{t}{\Delta t}\big(L(x,y;t+\Delta t) - L(x,y;t)\big) \quad (11)$$

where $\nabla^2 L(x,y;t)$ is the Laplacian of the Gaussian Operator defined in Laplacian of Gaussian (LoG) Blob detector. Fig. 18 shows DoG applied on the sample radiograph taken as input.

Determinant of Hessian (DoH) [36] is another fundamental method of Blob Detection, where the Monge-Ampère Operator is considered taking a scale-normalized determinant of Hessian and the whole algorithm can be mathematically denoted as:

$$(\hat{y}, \hat{w}; \hat{s}) = argmaxlocal_{(y,w;s)}((\det H_{norm} L)(y,w;s)) \quad (12)$$

Where $(\hat{y}, \hat{w})$ represents the blob points, $\hat{s}$ is the scale parameter and $HL$ is the Hessian Matrix on the scale-space orientation $L$. Fig. 18 represents the Determinant of Hessian (DoH) Blob Detector applied on the input sample radiograph.

*3) Feature Matching and Interest Points:*

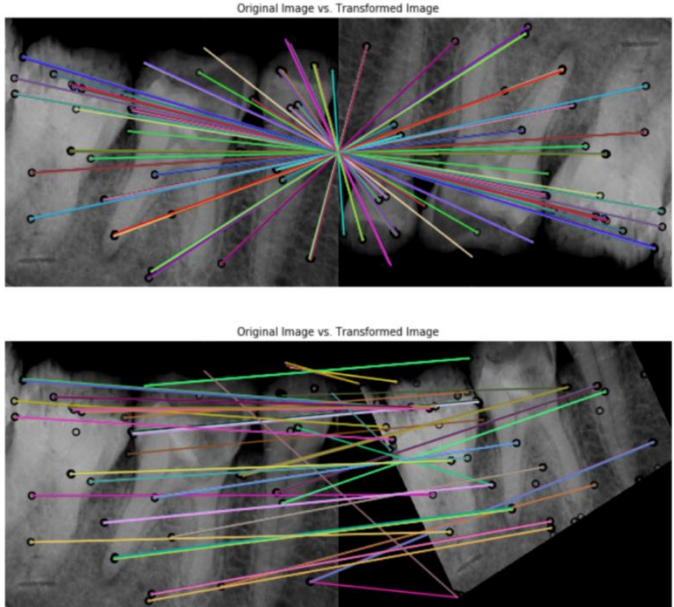

Fig. 19 (a). Feature Points Matching in Flipping Case (b) Feature Matching in Rotational Transformation Case

Feature Matching [37] is the process of finding and matching highly correlated feature points between the original image and the transformed image or finding the order of homography between two images. This is primarily done to for practical applications including image rectification, understanding rotational or translational transformation of images and also to rectify transformed images for image restoration and can be used in advanced applications like pose estimation or perspective correction. ORB (Oriented FAST and Rotated BRIEF) [38] feature detector and Binary descriptor was used to perform feature matching on the sample radiograph. ORB uses Hamming distance metric for feature matching while being invariant to changes in the scale or the rotational factors of the image. It was introduced as a better alternative to the SIFT key-point detector as it can be computed faster and was more robust and efficient. Fig. 19 shows how ORB and Binary descriptors were used to detect key-points and then for feature matching in cases of variance in the scale or the rotational factor. In radiography such as IOPA, feature matching can be used in 3-D reconstruction of the scanned oral cavity and to match feature markers in different perspective of view on the same region of diagnosis.

Interest Point detection is a process in computer vision which involves detection and processing of certain points or regions in the image which are well defined in terms of spatial orientation, mathematical definition and is an area of high information gain which can be symbolized by areas less susceptible to noise and having vivid texture patterns. Interest Point is predominately a preprocessing step for feature matching as the key-points detected in this procedure is used for feature matching to prove homography or can be used for 3-dimensional reconstruction of the image.

Fig. 20 shows the ORB Interest points detected in the sample radiograph and the high density key-points detected in this procedure was then used in the Feature Matching as shown in Fig 19.

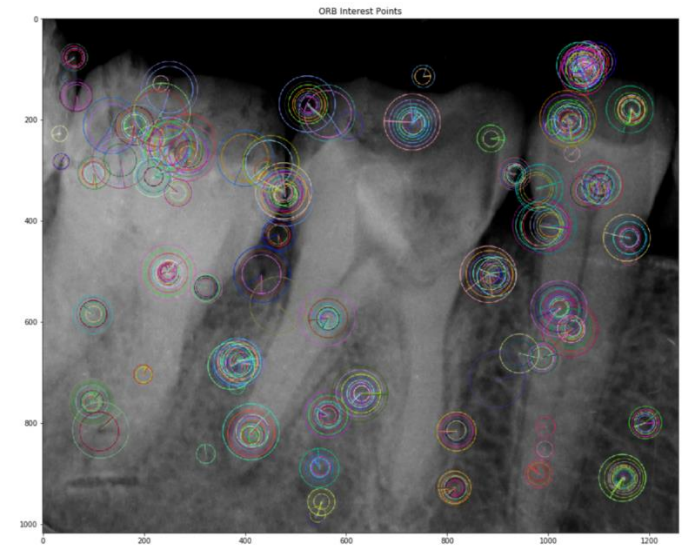

Fig. 20 Detected ORB Key-points on the sample radiograph of IOPA.

*4) Dense DAISY Descriptors:*

Dense DAISY descriptor [39] is a fast and efficient feature descriptor and detector used for dense matching. Similar to earlier conventional descriptors like SIFT (Scale Invariant and Feature Transform) and GLOH (Gradient Location and

Oriented Histogram), it is based on gradient oriented histograms but rather take up Gaussian weighting and symmetrically circular kernel which thus increases its speed and lowers the computation cost for dense feature matching. Feature Descriptors like DAISY can be used efficiently in feature matching, detecting feature points and also for reconstruction in case of radiography analysis in medical diagnosis. Fig. 21 shows the 42 extracted DAISY descriptors from the sample IOPA radiograph taken.

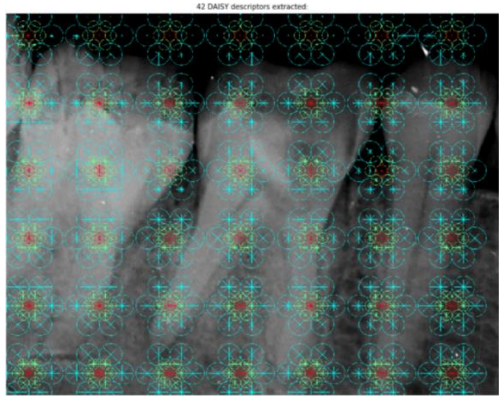

Fig. 21. Detected 42 Dense DAISY Descriptors on the sample IOPA radiograph.

*F. Edge Detection, Thresholding and Contour Modelling*

Edge Detection [40] is a high priority technique in any kind of Image Processing pipeline, it not only helps to establish outlining of edges in the image but also paves the way for contour modelling and object segmentation in the image. Over the years, researchers have come up with new robust edge detectors to provide more efficient and robust solution to this task. Edge can be defined as any point on the image where there is a sharp variation in pixel intensity or a region of high contrast separated by a visible boundary.

*1) Gabor, LMAK and PVMAK Edge Detectors:*

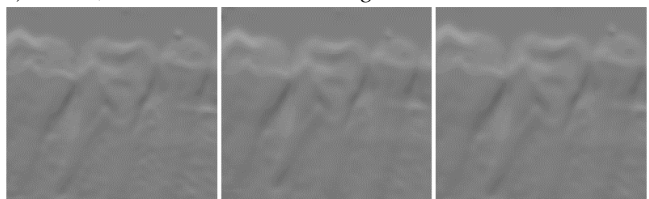

Fig. 22. (a) Gabor Edge Detector (b). LMAK Edge Detector (c) PVMAK Edge Detector.

Lower-level Edge Detectors are commonly used to establish early intuitions of the edges and patterns present in the image. Gabor Edge Detector is rather a very abstract and easy to implement Edge Detector. The mathematical foundation of Gabor is that of a Gaussian kernel modulated with a sine kernel. Fig. 22(a) shows the implementation of a Gabor Edge Detector on the sample radiograph taken.

LMAK (Lorentzian Modulated ArcTangent Kernel) and PVMAK (Pseudo-Voigt Modulated Arctangent Kernel) are similar abstract edge detectors which take inspiration from the Gaussian Gabor Edge Detector. Mathematically, LMAK and PVMAK use Lorentzian and Pseudo-Voigt Kernels instead of the Gaussian Kernel and modulate it with an Arctangent kernel rather than a sine one.

The mathematical equations for Voigt Function are shown in the subsequent equations:

$$V(x; \sigma, \gamma) = \int_{-\infty}^{\infty} G(x'; \sigma) L(x - x'; \gamma) \partial x' \quad (13)$$

$$L(x, y; \gamma) = \frac{\gamma}{\pi(x^2 + y^2)} \quad (14)$$

$$G(x; \sigma) = \frac{e^{-\frac{x^2}{2\sigma^2}}}{\sigma\sqrt{2\pi}} \quad (15)$$

$$V(x; \sigma, \gamma) = \frac{Re[w(z)]}{\sigma\sqrt{2\pi}} \quad (16)$$

$$z = x + \frac{iy}{\sigma\sqrt{2}} \quad (17)$$

The first equation represents the Voigt function, while the second and third equations are the centralized Lorentzian and Gaussian Functions respectively. The last two equations are the Real Integral part and the parameter z in respect to Fadeeva Equation respectively. However since Voigt Function is the convolution of Gaussian and Lorentzian Function, which is computationally extensive, we take the Pseudo-Voigt Profile which is represented as follows:

$$V_p(x) = \eta L(x, f) + (1 - \eta) G(x, f) \quad (18)$$

$$\eta = 1.36606 \left(\frac{f_L}{f}\right) - 0.47719 \left(\frac{f_L}{f}\right)^2 + 0.11116 \left(\frac{f_L}{f}\right)^3 \quad (19)$$

$$f = [f_G^5 + 2.69 f_G^4 f_L + 2.42 f_G^3 f_L^2 + 0.078 f_G f_L^4 + f_L^5]^{\frac{1}{5}} \quad (20)$$

$L(x, f)$ is the Lorentzian Function while $G(x, f)$ is the Gaussian function and $\eta$ is the factor of modulation, a parameter of Full-Width Half-Maximum(FWHM) values. $f_L$ and $f_G$ are the FWHM values of Lorentzian and Gaussian Distribution.

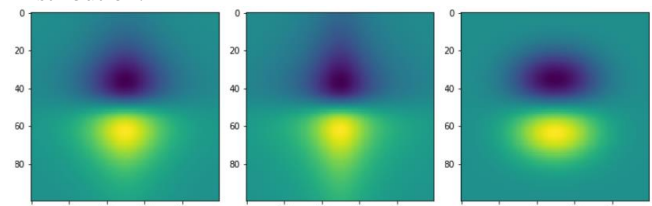

Fig. 23. (a) LMAK Kernel (b). PVMAK Kernel (c) Gaussian Gabor Kernel [22]

Fig 22(b) and 22(c) shows the LMAK and PVMAK Edge Detector applied on the input sample radiograph. The advantages of PVMAK and LMAK over Gabor is the smoothening factor of PVMAK and LMAK is higher and also they reduce sharpness thus noise while computing the edges.

Fig 23 shows the edge detector kernels of LMAK, PVMAK and Gaussian Gabor.

*2) Sobel, Roberts, Frangi, Hybrid Hessian, Spline and Scharr Filters and Thresholding:*

Frangi filter [41] and Hybrid Hessian filters [42] are used to compute continuous edges on the given image. Mostly based on volumetric object detection and edge mapping in images, Frangi and Hybrid Hessian filters are highly efficient in computing active edges and is highly popular in the case of medical image diagnosis especially in cases of detection of tube-like structures like blood vessels, tissues and fiber layers and can be applied thus in detection of anomalies in the region of diagnosis. Fig. 24 shows the application of Frangi and Hybrid Hessian Filters on the sample IOPA radiograph taken.

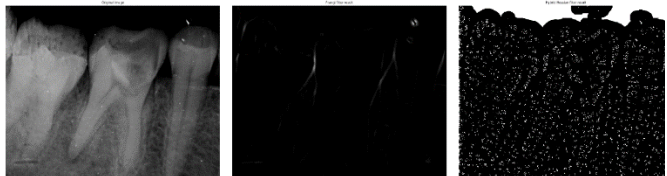

Fig. 24. (a) Original Sample IOPA Radiograph (b) Frangi Filter applied on the sample IOPA Radiograph (c) Hybrid Hessian Filter applied on the sample IOPA Radiograph.

Spline Filter [43] also takes inspiration from the Gaussian Filters for surface metrology and detection of edges. It overcomes the shortcomings imposed by Gaussian Filters due to irregularity in edges or discontinuous texture patterns. Spline filters can be extremely useful in detecting pulp infection in teeth evident on an oral cavity radiograph. Fig. 25 shows the application of Spline Filter on the sample IOPA Radiograph taken.

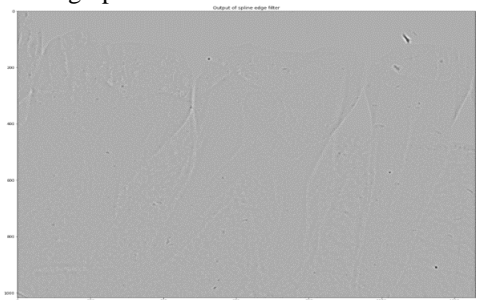

Fig. 25. Spline Filter applied on the sample IOPA Radiograph.

Sobel filters [44] are one of the most important edge detection algorithms presents which compute edges both on the horizontal and the vertical orientation. Sobel operates using two kernels for computing the horizontal and vertical edges first and then uses the following mathematical notation to approximate both the gradients computed:
$$G = \sqrt{G_x^2 + G_y^2} \tag{21}$$
, where $G_x^2$ and $G_y^2$ are the squares of the gradient in horizontal orientation and in vertical orientation. The gradient's direction is the computed using the formula:
$$\ominus = \operatorname{atan}(\frac{G_x}{G_y}). \tag{22}$$

Fig. 26 shows the application of the Sobel Filter on individual RGB channels of the image and the Value Converted Image (HSV) of the sample IOPA radiograph based on variation in Hue, Saturation and Lighting intensity functions of that radiograph.

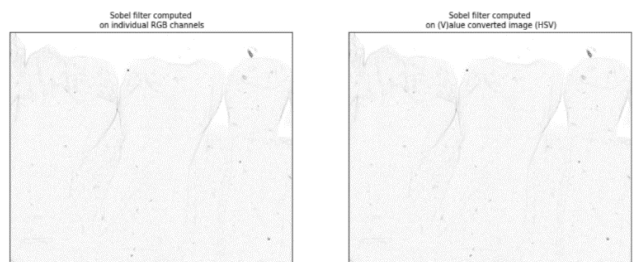

Fig. 26. (a) Sobel Filter computed on Individual RGB Channels. (b). Sobel Filter computed on the Value Converted Image based on their HSV (Hue, Saturation and Lightness values)..

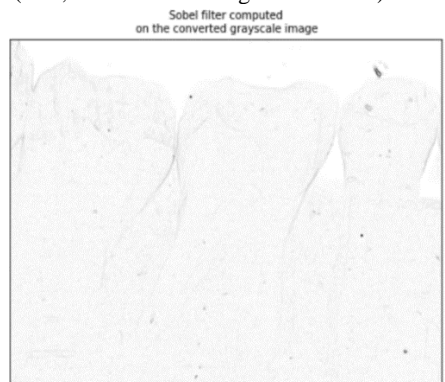

Fig. 27. Sobel Filter applied on the sample grayscale converted IOPA Radiograph.

Fig. 27 shows the Sobel filter applied on the grayscale converted sample radiograph taken while Fig 28 shows both the Horizontal and Vertical Edges detected by Sobel Y and Sobel X filters respectively.

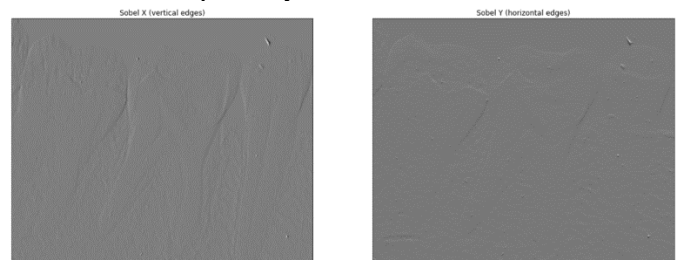

Fig. 28.(a)Vertical Edges Detected by Sobel X filter. (b)Horizontal Edges detected by Sobel Y Filter.

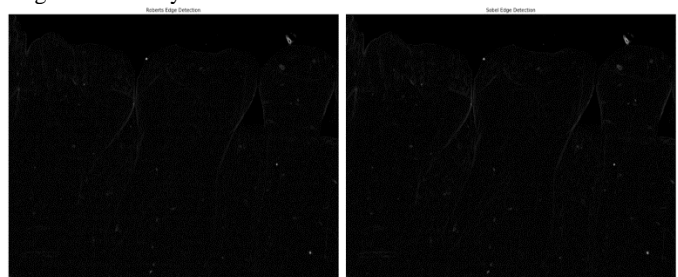

Fig. 29.(a) Roberts Edge Detector applied on the sample IOPA radiograph. (b). Sobel Filter applied on the sample IOPA Radiograph.

Fig 29 shows the comparison between the Roberts Edge Detector and Sobel Edge Detector. Roberts Edge Detector is another robust edge detector similar to Sobel Edge Detector with a slight change in gradient's direction mathematical representation which is as follows:

$$\ominus = \arctan\left(\frac{G_x}{G_y}\right) - \frac{3\pi}{4} \quad (23)$$

Roberts Edge Detector has the following foundational principles for edge computation: the edges computed should be vivid and sharp while the background should involve the minimum noise and the intensities of the edges computed should be at par with human-level of perceiving edges.

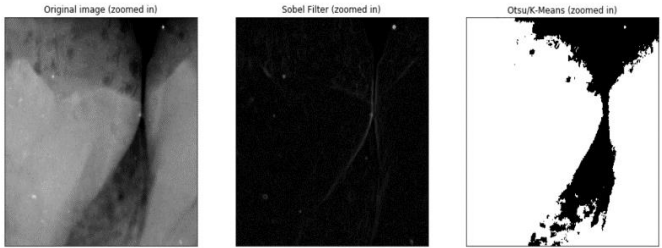

Fig. 30.(a) Zoomed in original IOPA radiograph. (b). Sobel Filter applied on the sample zoomed IOPA Radiograph. (c). Otsu/ K-Means Masked Segmentation of the sample zoomed IOPA Radiograph.

Fig 30 shows the comparison between the Sobel Filter applied on the zoomed in area of the sample IOPA radiograph and the same area's effective segmentation using K-Means involving an Otsu Threshold mask [45]. Threshold is an important aspect in Image Processing as it helps in the separation/ segmentation of objects present in the image. Fig 31 shows basic threshold while Fig 32 shows the comparison between various methods of Threshold including masked threshold and K-Means segmentation [46].

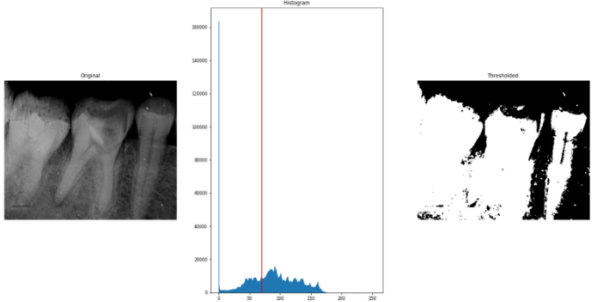

Fig. 31.(a) Original Sample IOPA Radiograph. (b). Thresholded Value shown in the pixel intensity graph. (c) Thresholded Image.

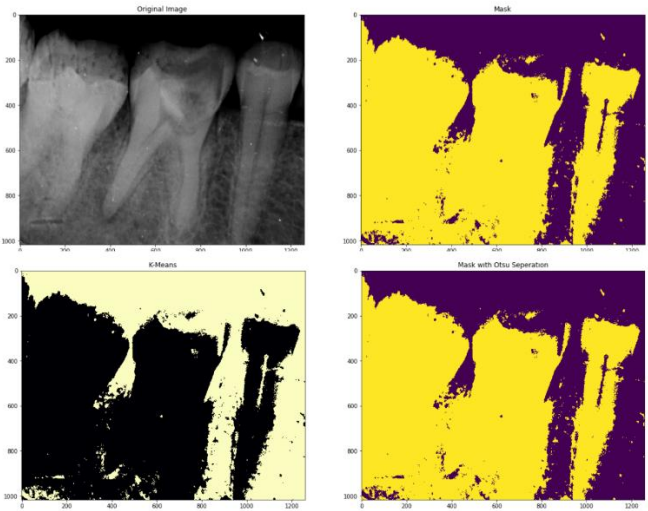

Fig. 32.(a) Original Sample IOPA Radiograph. (b). Mask Thresholded Image. (c) K-Means Segmented/Thresholded Image. (d). Masked with Otsu separation thresholded image.

Thresholding is flexible and the thresholded value can be changed to produce better results for efficient segmentation as shown in Fig 33.

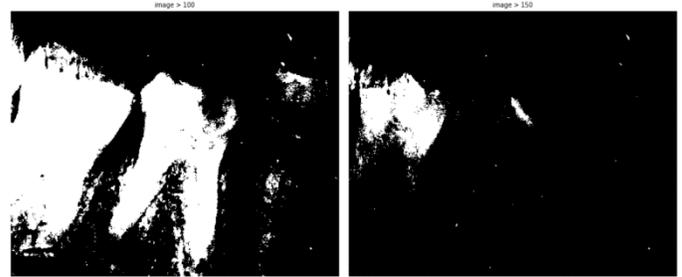

Fig. 33.(a)Threshold Value set to 100. (b). Threshold Value set 150.

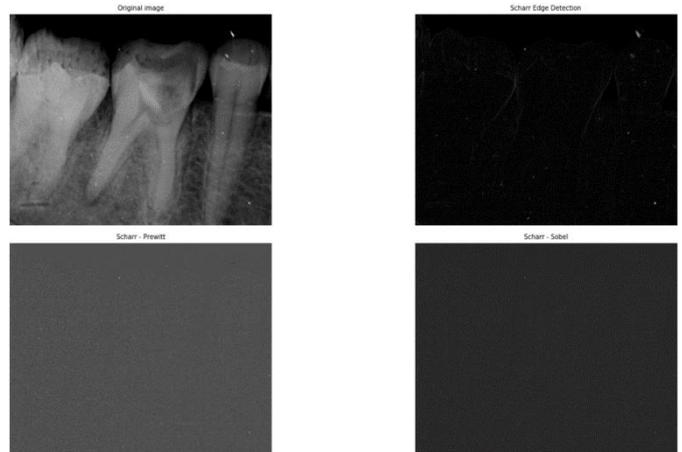

Fig. 34.(a) Original Sample IOPA Radiograph. (b).Scharr Edge Detector (c)Scharr-Prewitt Edge Detector. (d).Scharr-Sobel Edge Detector.

Fig 34 shows the comparative analysis of various modulated forms of Scharr Edge Detectors [47] with the original Scharr Edge Detector. Scharr Edge Detectors are used for computing vertical and horizontal edges based on the Scharr Transform and prove to be an efficient alternative to conventional Prewitt or Sobel horizontal and vertical edge filters.

*3) Canny Edge Detection and Contour Mapping:*

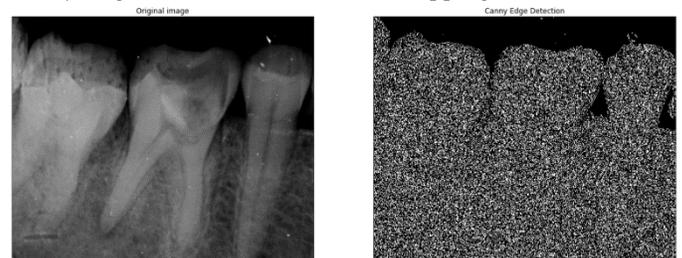

Fig. 35.(a) Original Sample IOPA Radiograph. (b).Masked Canny Edge Detector.

Canny Edge Detector [48] is a highly efficient and robust edge detector used in edge detection and contour mapping. Although the Canny filter is a great, simple yet precise edge detection algorithm, it can be highly modulated and modified to reduce its drawbacks while computing the edges in a more robust fashion. One of the ways is to use smoothened Canny

Masked with Otsu filters for edge detection. Fig 35 shows the application of a masked canny operator on the sample radiograph while Fig 36 shows the comparison of filter values between canny and smoothened canny with Otsu threshold value. Subsequently Fig. 37 shows the application of that smoothened canny mask on the sample IOPA Radiograph taken.

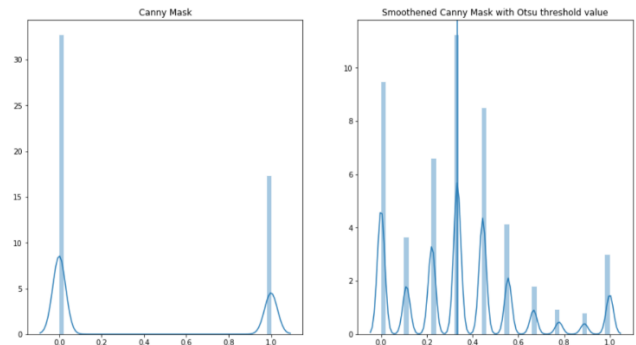

Fig. 36.(a)Canny Mask (b).Masked Smoothened Canny with Otsu Threshold Value.

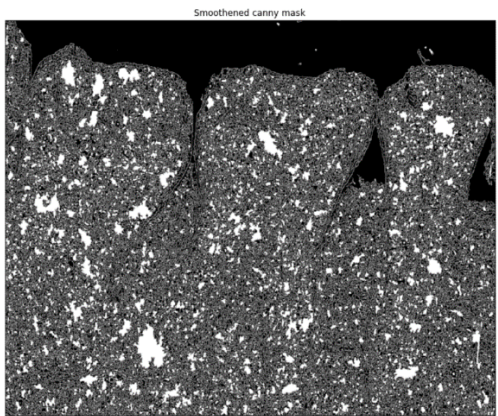

Fig. 37. Masked Smoothened Canny with Otsu Threshold Value applied on the Sample IOPA Radiograph.

Contour Mapping is the subsequent process in edge detection and is used to define boundaries between objects present in the image as a preprocessing step for edge detection. Fig. 38 and 39 show contour modelling both on the original IOPA radiograph and in the Cartesian co-ordinate system while Fig 40 shows contour modelling using colored Canny Filter.

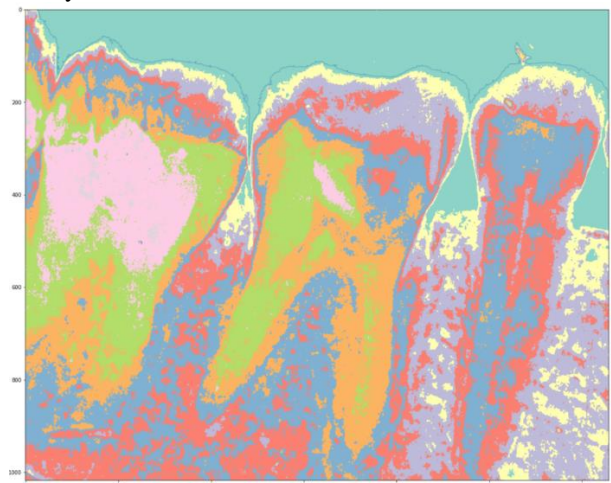

Fig. 38. Contour Mapping on the Original Sample Radiograph.

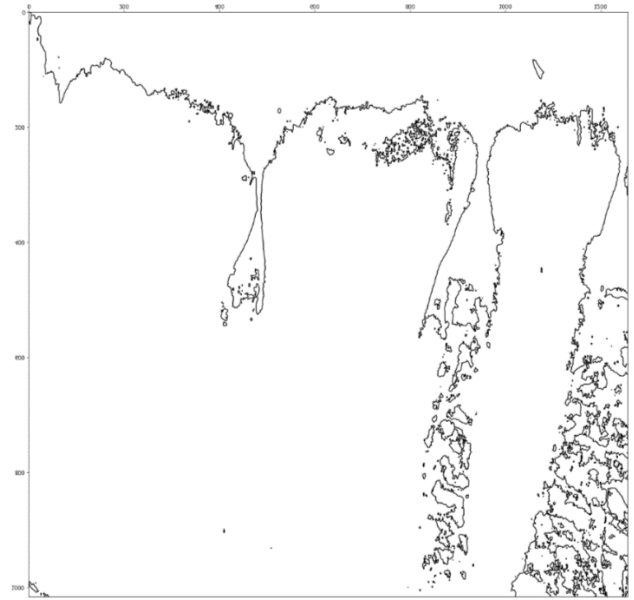

Fig. 39. Contour Mapping on the Original Sample Radiograph in Cartesian Co-Ordinate System.

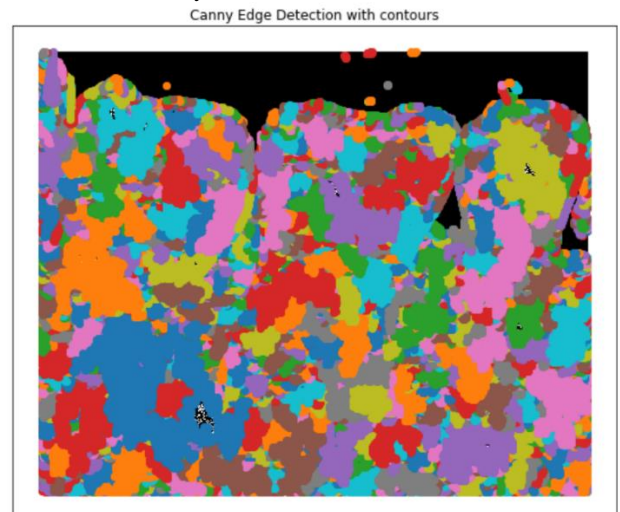

Fig. 40.Colored Canny Edge Detection with Contour Mapping on the sample IOPA Radiograph.

*G. Segmentation*

Segmentation of Objects in an image has always been an important task in the Image Processing pipeline. Segmentation is a high priority task in medical image processing but remains still a difficult task to perfect due to high variation in images. Segmentation divides the image into meaningful areas to analyze. Image Segmentation is a post edge detection and contour mapping process used to locate objects and also to give labels to objects. Image Segmentation [49] in the medical science domain has a lot of end-to-end practical as well as research applications including pathological diagnosis or detecting tumors or cancerous tissues, for analyzing volumetric tissues, for surgery planning, et cetera.

Fig. 41 shows effective watershed segmentation algorithm, applied on the sample IOPA Radiograph taken.

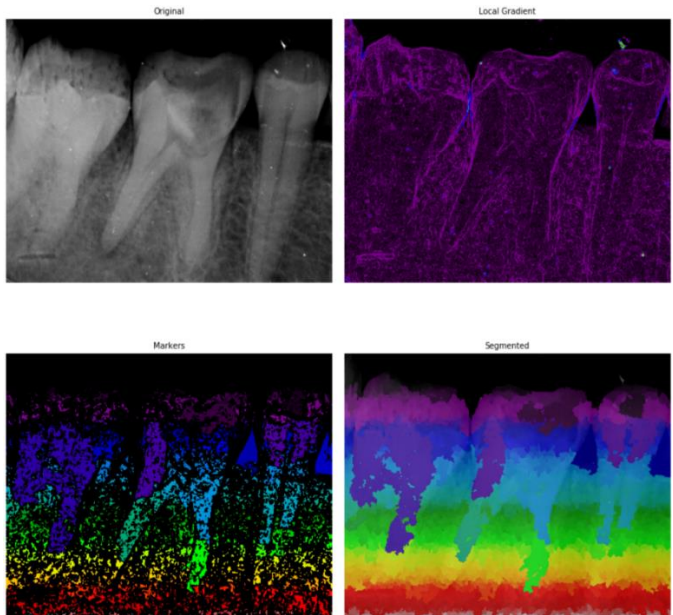

Fig. 41. (a). Original Sample IOPA Radiograph. (b). Local Gradients Computed on the Sample IOPA Radiograph. (c). Obtained Markers for Segmentation. (d). Watershed Segmentation applied on the sample IOPA Radiograph.

Watershed Algorithm [46] is a classic segmentation algorithm used in Image Processing. It follows a strict procedure of segmenting the image based on the markers obtained which are computed based on the area of low gradient value in the image. Technically, an area of high gradient in the image defines the boundaries separating the objects present in the image.

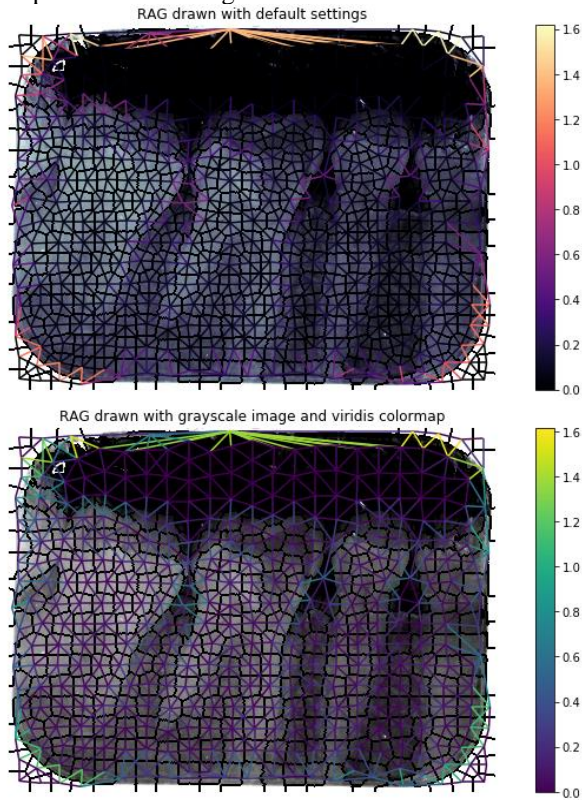

Fig. 42. (a). RAG on the original IOPA Sample Radiograph. (b). RAG on the grayscale converted IOPA Sample Radiograph.

RAG (Region Adjacency Graph) [53] is another popular method of image segmentation. It uses an interactive graphical representation of the segmented regions in the image where each region of the image can be represented as a node in the graph and the boundary is set between every adjacent region. The regions can be defined using certain weights, which here are computed based on their average color and intensity values mapped by the pixels in that region. Fig. 42 shows the application of RAG on the sample IOPA Radiograph taken.

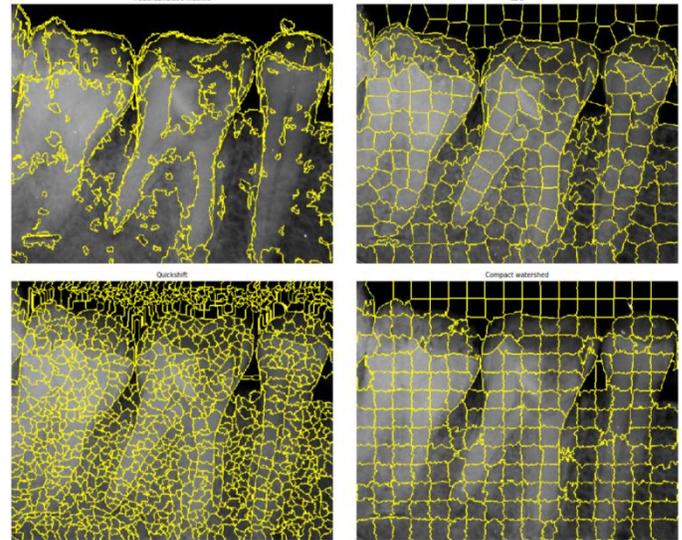

Fig. 43. (a) .Felzenszwalb's efficient graph based segmentation applied on the sample IOPA Radiograph. (b) SLIC - K-Means based image segmentation applied on the sample IOPA Radiograph. (c).Quickshift image segmentation applied on the sample IOPA Radiograph. (d).Compact watershed segmentation of gradient images applied on the sample IOPA Radiograph.

Fig 43 shows the comparative analysis of the various standard low-level super-pixel image segmentation algorithms widely used in various image processing techniques. These algorithms include: Felzenszwalb's efficient graph based segmentation, SLIC- K-Means based image segmentation, Quickshift Segmentation and Compact Watershed Segmentation of Gradient Images.

Felzenszwalb's efficient graph-based segmentation [50] involves a fast 2-dimensional segmentation algorithm which has a varying scale parameter that can be used to vary the segmented region's size based on the graph representation of the image which is used to define the boundaries defining the segments in the image.

SLIC- K-Means based image segmentation [52] is considered to be a standard image segmentation algorithm as it involves K-Means clustering which is easier to compute based on 5-dimensional space of color information and the image. As seen in Fig 43, SLIC is highly efficient in perfectly carving and defining the boundaries or edges defining the root of the teeth and can be effectively used in clinical research for the study of root erosion or IOPA lesions in oral pathology.

Quickshift Segmentation [51] is a novel approach towards 2-dimensional segmentation of objects in image. It uses an approximation of kernelized mean shift belonging to local mode seeking algorithms. It incorporates computing a hierarchical segmentation model on various scales simultaneously.

Compact Watershed Segmentation of Gradient Images as described earlier takes the local gradients mapped by the grayscale version of the image and decides markers based on regions of low gradient values where high gradient values define the boundaries between segments.

## V. Experimental set-up and observations

The research was able to build a comprehensive knowledge of how various image processing techniques can be used for achieving optimal diagnostic results and accelerating the process of identifying pathologies, increasing robustness to decrease cases of False Positive and True Negative cases in Diagnostic reports of Oral Pathologies, pertaining to IOPA radiographs focusing on Apical Periodontitis, a common pathology affecting thousands. It also provides a descriptive analytical approach towards understanding IOPA radiographs and how Image Processing techniques can be applied for advancing clinical research in Biomedical Image Processing.

The research case study was constructed on a sample IOPA-Apical Periodontitis and Dental Caries case radiograph obtained from a male subject, aged 19. The machine specification used to obtain the radiograph along with the specification of the machine used for performing the image processing analysis on the radiograph is as follows:

TABLE I
EXPERIMENTAL SETUP

| Machine Name | Specification (Model Number, Internal Specifications) |
|---|---|
| Intra-Oral Dental X-Ray Machine | Timex-70 |
| MSI GP-63 | Leonard 8RE, GTX 1060 GPU, Intel Core i-7 Processor |

Python programming language for performing all the simulations and the image processing on the sample radiograph. The sub-modules used involved Scikit-Image, Numpy, Seaborn, OpenCV and Matplotlib.

## VI. Conclusion

The research is aimed at providing academia and clinical pathological experts a comprehensive guide of performing advanced diagnosis and accelerating the diagnostic pipeline by involving active image processing techniques. With advanced Image Processing techniques like Noise Removal, Segmentation, Feature Extraction and Contour Modelling, dental professionals can advance and improve the diagnosis of Apical Periodontitis by using these techniques while analyzing IOPA Radiographs.

Future Scope of the research includes building of more efficient image processing pipelines to address all six lesions of IOPA which includes- Apical Periodontitis, Condensing Osteitis, Periapical Abscess, Periapical Cyst, Periapical Granuloma and Rarefying Osteitis. Future work also includes construction of a mobile software based diagnostic pipeline of IOPA radiograph diagnosis by using advanced computer vision techniques for automatic segmentation, classification and localization of pathologies present in the radiograph.

## VII. References

[1] Lo, Winnie Y., and Sarah M. Puchalski. "Digital image processing." Veterinary Radiology & Ultrasound 49 (2008): S42-S47.
[2] Jayachandran, Sadaksharam. "Digital imaging in dentistry: A review" Contemporary clinical dentistry 8, no. 2 (2017): 193.
[3] Ghom, Anil Govindrao. Basic oral radiology. JP Medical Ltd, 2014, pp. 1-5
[4] White, Stuart C., and Michael J. Pharoah. Oral radiology-E-Book: Principles and interpretation. Elsevier Health Sciences, 2014.
[5] Gupta, A., P. Devi, R. Srivastava, and B. Jyoti. "Intra oral periapical radiography-basics yet intrigue: A review." Bangladesh Journal of Dental Research & Education 4, no. 2 (2014): 83-87.
[6] Van Der Stelt, Paul F. "Filmless imaging: the uses of digital radiography in dental practice." The Journal of the American Dental Association 136, no. 10 (2005): 1379-1387.
[7] Baghdady, Mariam T. "Principles of radiographic interpretation." White and Pharoah's Oral Radiology E-Book: Principles and Interpretation (2018): 290.
[8] Reddy, MV Bramhananda, Varadala Sridhar, and M. Nagendra. "Dental x-ray image analysis by using image processing techniques". International Journal of Advanced Research in Computer Science and Software Engineering 2, no. 6 (2012).
[9] Oprea, Stefan, Costin Marinescu, Ioan Lita, Mariana Jurianu, Daniel Alexandru Visan, and Ion Bogdan Cioc. "Image processing techniques used for dental x-ray image analysis". In Electronics Technology, 2008. ISSE'08. 31st International Spring Seminar on, pp. 125-129. IEEE, 2008.
[10] Leo, L. Megalan, and T. Kalpalatha Reddy. "Digital Image Analysis in Dental Radiography for Dental Implant Assessment: Survey". International Journal of Applied Engineering Research 9, no. 21 (2014): 10671-10680.
[11] Patil, Harshada, and Sagar A. More. "Dental Image Processing In Endodontic Treatment" (2017).
[12] Deserno, Thomas M. "Fundamentals of biomedical image processing". In Biomedical Image Processing, pp.1-51. Springer, Berlin, Heidelberg, 2010.
[13] McAuliffe, Matthew J., Francois M. Lalonde, Delia McGarry, William Gandler, Karl Csaky, and Benes L.Trus. "Medical image processing, analysis and visualization in clinical research". In Computer-Based Medical Systems, 2001. CBMS 2001. Proceedings. 14th IEEE Symposium on, pp. 381-386. IEEE, 2001.
[14] Ronneberger, Olaf, Philipp Fischer, and Thomas Brox. "U-net: Convolutional networks for biomedical image segmentation". In International Conference on Medical image computing and computer-assisted intervention, pp. 234-241. Springer, Cham, 2015.
[15] Dhawan, Atam P. "A review on biomedical image processing and future trends". Computer Methods and Programs in Biomedicine 31, no. 3-4 (1990): 141-183.
[16] Newman, Michael G., Henry Takei, Perry R. Klokkevold, and Fermin A. Carranza. "Carranza's clinical periodontology". Elsevier health sciences, 2011.
[17] Rajendran, R. Shafer's textbook of oral pathology. Elsevier India, 2009, pp.669-676
[18] Newman, Michael G., Henri H. Takei, Fermin A. Carranza, and Saunders WB. "Clinical Periodontology" (2006).
[19] Heitz-Mayfield, Lisa JA, Marc Schätzle, Harald Löe, Walter Bürgin, Åge Ånerud, Hans Boysen, and Niklaus P. Lang. "Clinical course of chronic periodontitis: II. Incidence, characteristics and time of occurrence of the initial periodontal lesion". Journal of clinical periodontology 30, no. 10 (2003): 902-908.
[20] Ghom, Anil Govindrao, and Savita Anil Lodam Ghom, eds. Textbook of oral medicine. JP Medical Ltd, 2014 pp. 554-556
[21] Kang, G. "Digital image processing." Quest, vol. 1, Autumn 1977, p. 2-20. 1 (1977): 2-20.

[22] Diganta Misra, "Robust Edge Detection using Pseudo Voigt and Lorentzian modulated arctangent kernel,"8th IEEE International Advanced Computing Conference ,2018 (IACC)., to be published.
[23] Damon, James. "Generic structure of two-dimensional images under Gaussian blurring." SIAM Journal on Applied Mathematics 59, no. 1 (1998): 97-138.
[24] Gupta, Gajanand. "Algorithm for image processing using improved median filter and comparison of mean, median and improved median filter." International Journal of Soft Computing and Engineering (IJSCE) 1, no. 5 (2011): 304-311.
[25] Tomasi, Carlo, and Roberto Manduchi. "Bilateral filtering for gray and color images." In Computer Vision, 1998. Sixth International Conference on, pp. 839-846. IEEE, 1998.
[26] Dougherty, Edward R., and Roberto A. Lotufo. Hands-on morphological image processing. Vol. 59. SPIE press, 2003.
[27] Yu-qian, Zhao, Gui Wei-hua, Chen Zhen-cheng, Tang Jing-tian, and Li Ling-Yun. "Medical images edge detection based on mathematical morphology." In Engineering in Medicine and Biology Society, 2005. IEEE-EMBS 2005. 27th Annual International Conference of the, pp. 6492-6495. IEEE, 2006.
[28] Vincent, Luc. "Morphological grayscale reconstruction in image analysis: applications and efficient algorithms." IEEE transactions on image processing 2, no. 2 (1993): 176-201.
[29] Vincent, Luc. "Morphological area openings and closings for grey-scale images." In Shape in Picture, pp. 197-208. Springer, Berlin, Heidelberg, 1994.
[30] Evans, Adrian N., and X. U. Lin. "A morphological gradient approach to color edge detection." IEEE Transactions on Image Processing 15, no. 6 (2006): 1454-1463.
[31] Vasantha, M., V. Subbiah Bharathi, and R. Dhamodharan. "Medical image feature, extraction, selection and classification." International Journal of Engineering Science and Technology 2, no. 6 (2010): 2071-2076.
[32] Zhu, Qiang, Mei-Chen Yeh, Kwang-Ting Cheng, and Shai Avidan. "Fast human detection using a cascade of histograms of oriented gradients." In Computer Vision and Pattern Recognition, 2006 IEEE Computer Society Conference on, vol. 2, pp. 1491-1498. IEEE, 2006.
[33] Lindeberg, Tony. "Detecting salient blob-like image structures and their scales with a scale-space primal sketch: A method for focus-of-attention." International Journal of Computer Vision 11, no. 3 (1993): 283-318.
[34] Sotak Jr, George E., and Kim L. Boyer. "The Laplacian-of-Gaussian kernel: a formal analysis and design procedure for fast, accurate convolution and full-frame output." Computer Vision, Graphics, and Image Processing 48, no. 2 (1989): 147-189.
[35] Wang, Shoujia, Wenhui Li, Ying Wang, Yuanyuan Jiang, Shan Jiang, and Ruilin Zhao. "An Improved Difference of Gaussian Filter in Face Recognition." Journal of Multimedia 7, no. 6 (2012): 429-433.
[36] Froese, Brittany D., and Adam M. Oberman. "Convergent finite difference solvers for viscosity solutions of the elliptic Monge–Ampère equation in dimensions two and higher." SIAM Journal on Numerical Analysis 49, no. 4 (2011): 1692-1714.
[37] Mian, Ajmal S., Mohammed Bennamoun, and Robyn A. Owens. "A novel representation and feature matching algorithm for automatic pairwise registration of range images." International Journal of Computer Vision 66, no. 1 (2006): 19-40.
[38] Abdel-Hakim, Alaa E., and Aly A. Farag. "CSIFT: A SIFT descriptor with color invariant characteristics." In Computer Vision and Pattern Recognition, 2006 IEEE Computer Society Conference on, vol. 2, pp. 1978-1983. IEEE, 2006.
[39] Tola, Engin, Vincent Lepetit, and Pascal Fua. "Daisy: An efficient dense descriptor applied to wide-baseline stereo." IEEE transactions on pattern analysis and machine intelligence 32, no. 5 (2010): 815-830.
[40] Kumar, Indrajeet, Jyoti Rawat, and H. S. Bhadauria.," A conventional study of edge detection technique in digital image processing",International Journal of Computer Science and Mobile Computing 3, no. 4 (2014): pp. 328-334.
[41] Jimenez-Carretero, Daniel, Andres Santos, Sjoerd Kerkstra, Rina Dewi Rudyanto, and Maria J. Ledesma-Carbayo. "3D Frangi-based lung vessel enhancement filter penalizing airways." In Biomedical imaging (ISBI), 2013 IEEE 10th international symposium on, pp. 926-929. IEEE, 2013.
[42] Ng, Choon-Ching, Moi Hoon Yap, Nicholas Costen, and Baihua Li. "Automatic wrinkle detection using hybrid Hessian filter." In Asian Conference on Computer Vision, pp. 609-622. Springer, Cham, 2014.
[43] Goto, Tomonori, Jota Miyakura, Kozo Umeda, Soichi Kadowaki, and Kazhisa Yanagi. "A robust spline filter on the basis of L2-norm." Precision engineering 29, no. 2 (2005): 157-161.
[44] Aqrawi, Ahmed Adnan, and Trond Hellem Boe. "Improved fault segmentation using a dip guided and modified 3D Sobel filter." In SEG Technical Program Expanded Abstracts 2011, pp. 999-1003. Society of Exploration Geophysicists, 2011.
[45] Liu, Chen-Chung, Chung-Yen Tsai, Jui Liu, Chun-Yuan Yu, and Shyr-Shen Yu. "A pectoral muscle segmentation algorithm for digital mammograms using Otsu thresholding and multiple regression analysis." Computers & Mathematics with Applications 64, no. 5 (2012): 1100-1107.
[46] Ng, H. P., S. H. Ong, K. W. C. Foong, P. S. Goh, and W. L. Nowinski. "Medical image segmentation using k-means clustering and improved watershed algorithm." In Image Analysis and Interpretation, 2006 IEEE Southwest Symposium on, pp. 61-65. IEEE, 2006.
[47] AlNouri, Mason, Jasim Al Saei, Manaf Younis, Fadi Bouri, Mohamed Ali Al Habash, Mohammed Hamza Shah, and Mohammed Al Dosari. "Comparison of Edge Detection Algorithms for Automated Radiographic Measurement of the Carrying Angle." Journal of Biomedical Engineering and Medical Imaging 2, no. 6 (2016): 78.
[48] Green, Bill. "Canny edge detection tutorial." Retrieved: March6 (2002): 2005.
[49] Pham, Dzung L., Chenyang Xu, and Jerry L. Prince. "Current methods in medical image segmentation." Annual review of biomedical engineering 2, no. 1 (2000): 315-337.
[50] Felzenszwalb, Pedro F., and Daniel P. Huttenlocher. "Efficient graph-based image segmentation." International journal of computer vision 59, no. 2 (2004): 167-181.
[51] Vedaldi, Andrea, and Stefano Soatto. "Quick shift and kernel methods for mode seeking." In European Conference on Computer Vision, pp. 705-718. Springer, Berlin, Heidelberg, 2008.
[52] Achanta, Radhakrishna, Appu Shaji, Kevin Smith, Aurelien Lucchi, Pascal Fua, and Sabine Süsstrunk. "SLIC superpixels compared to state-of-the-art superpixel methods." IEEE transactions on pattern analysis and machine intelligence 34, no. 11 (2012): 2274-2282.
[53] Géraud, Thierry, J-F. Mangin, Isabelle Bloch, and Henri Maître. "Segmenting internal structures in 3D MR images of the brain by Markovian relaxation on a watershed based adjacency graph." In Image Processing, 1995. Proceedings., International Conference on, vol. 3, pp. 548-551. IEEE, 1995.